\documentclass{article}

\usepackage[nonatbib, preprint]{neurips_2025}

\usepackage[utf8]{inputenc} 
\usepackage{fontenc}    
\usepackage{hyperref}       
\usepackage{url}            
\usepackage{booktabs}       
\usepackage{amsfonts}       
\usepackage{nicefrac}       
\usepackage{microtype}      
\usepackage{graphicx}
\usepackage{amsmath}
\usepackage{amssymb}
\usepackage{tikz}
\usepackage{subcaption}
\usetikzlibrary{positioning, shapes.geometric, arrows.meta}

\title{Disentangling Polysemantic Neurons with a Null-Calibrated Polysemanticity Index and Causal Patch Interventions}

\author{
  Manan Gupta, Dhruv Kumar \\
  BITS Pilani, India \\
  \texttt{f20241231@pilani.bits-pilani.ac.in} \\
  \texttt{manangupta9997@gmail.com} \\
  \texttt{dhruv.kumar@pilani.bits-pilani.ac.in} \\
}

\begin{document}

\maketitle

\begin{abstract}
Neural networks often contain polysemantic neurons that respond to multiple, sometimes unrelated, features, complicating mechanistic interpretability. We introduce the Polysemanticity Index (PSI), a null-calibrated metric that quantifies when a neuron's top activations decompose into semantically distinct clusters. PSI multiplies three independently calibrated components: geometric cluster quality (S), alignment to labeled categories (Q), and open-vocabulary semantic distinctness via CLIP (D). On a pretrained ResNet-50 evaluated with Tiny-ImageNet images, PSI identifies neurons whose activation sets split into coherent, nameable prototypes, and reveals strong depth trends: later layers exhibit substantially higher PSI than earlier layers. We validate our approach with robustness checks (varying hyperparameters, random seeds, and cross-encoder text heads), breadth analyses (comparing class-only vs. open-vocabulary concepts), and causal patch-swap interventions. In particular, aligned patch replacements increase target-neuron activation significantly more than non-aligned, random, shuffled-position, or ablate-elsewhere controls. PSI thus offers a principled and practical lever for discovering, quantifying, and studying polysemantic units in neural networks.
\end{abstract}

\section{Introduction}

The increasing use of deep neural networks in high-stakes fields, such as medical diagnostics and autonomous navigation, has made model interpretability crucial. Understanding these complex models is vital for debugging, ensuring fairness, and building trust. A common approach is to analyze individual neurons as detectors for specific features. However, this is complicated by polysemanticity, where one neuron can respond to a mix of seemingly unrelated concepts \cite{elhage2022toy}. Similar to how a word can have multiple meanings based on context, a single neuron may activate for diverse visual features, like cat faces and car fronts. This ambiguity is a significant hurdle to achieving mechanistic interpretability of neural networks.

Although powerful, current interpretability paradigms have limitations in tackling this challenge. Feature visualisation techniques were the first to provide compelling, though qualitative, evidence of polysemanticity. These techniques generated synthetic images that maximally activated individual neurons, often resulting in a collage of unrelated visual elements \cite{erhan2009visualizing, nguyen2016multifaceted, olah2017feature}. These methods function like an observational science, powerfully demonstrating the existence of a phenomenon but lacking a framework for its quantification. On the other hand, quantitative concept-alignment methods, such as Network Dissection  \cite{bau2017network} or Testing with Concept Activation Vectors (TCAV) \cite{kim2018interpretability}, typically evaluate whether a neuron's activity aligns with a single, pre-defined human concept. They are designed to answer hypothesis-driven questions like "how important is the concept of 'stripes' for classifying a 'zebra'?" rather than asking the exploratory question of whether a neuron cleanly represents multiple distinct concepts simultaneously.

This methodological gap between exploratory, qualitative observation and rigid, quantitative hypothesis testing leaves a critical question unanswered: how can one systematically identify which neurons are polysemantic and then characterize the multiple concepts they encode? This work introduces the \textbf{Polysemanticity Index (PSI)}, a composite, quantitative, and statistically-grounded metric explicitly designed to bridge this gap. PSI is built upon three pillars of evidence, each probing a different facet of a neuron's response properties. For a given neuron, we first mine the image patches that cause its highest activations. We then assess: (i) the geometric separability of these patches in a semantic embedding space, quantifying whether they form distinct clusters; (ii) the class-label alignment of these clusters, measuring if they correspond to different ground-truth data categories; and (iii) the open-vocabulary semantic distinctness of the clusters, using a multimodal model (CLIP) to determine if each cluster can be assigned a unique, unambiguous textual concept \cite{radford2021learning}.

Our methodology includes rigorous null calibration for each component. By comparing raw scores to a carefully constructed null distribution, we derive statistical significance scores that prevent spurious findings and confirm that a high PSI score indicates a meaningful neuronal response. This approach starts with observed neuronal behavior to discover concepts post hoc, then provides a robust statistical score to quantify the strength of these conceptual decompositions.

We apply this framework to analyze the channels of a pretrained ResNet-50 model on the Tiny-ImageNet dataset \cite{he2016deep, le2015tiny}. Our empirical investigation yields several key findings. First, PSI robustly distinguishes real neuronal structure from chance, achieving an Area Under the Receiver Operating Characteristic (AUROC) of approximately 0.99 when separating late-layer neurons from a null baseline. Second, we uncover a striking hierarchical pattern: polysemanticity appears to be an emergent property of network depth, with neurons in \texttt{layer4} exhibiting significantly higher PSI scores than those in \texttt{layer3}. Third, we demonstrate that the feature prototypes discovered by our method are not merely correlational but are causally relevant to the neuron's function. Through a series of targeted patch-swap interventions, we show that inserting a patch aligned with a discovered prototype significantly boosts the neuron's activation compared to various controls.

In summary, our contributions are: (1) a novel, null-calibrated Polysemanticity Index (PSI) that combines geometric, categorical, and open-vocabulary semantic evidence; (2) a comprehensive interpretability pipeline, including causal interventions, for validating polysemantic structure; and (3) an empirical study revealing depth-wise patterns of polysemanticity in a widely-used CNN architecture, providing a new tool for the mechanistic analysis of neural representations.

\section{Related Work}


\paragraph{Neuron Interpretability and Polysemanticity.}

Early efforts to interpret individual neurons focused on their preferred stimuli. \textbf{Feature Visualization} techniques, developed by Erhan et al. \cite{erhan2009visualizing} and refined by Nguyen et al. \cite{nguyen2016multifaceted} and Olah et al. \cite{olah2017feature}, utilize optimization to create synthetic input images that maximally activate specific neurons. These methods provided compelling evidence of polysemanticity, as the "optimal" images often displayed a mix of unrelated objects, textures, and patterns. However, while these approaches are foundational for understanding neuronal function, they remain primarily qualitative and do not furnish a quantitative assessment of polysemanticity or its statistical significance.

Seeking a more quantitative approach, \textbf{Network Dissection} \cite{bau2017network} proposed a framework to automatically associate individual convolutional units with human-interpretable concepts. By measuring the spatial agreement between a unit's activation map and a large dictionary of semantic segmentation masks (e.g., for objects, parts, materials), this method assigns a concept label (like "dog detector") to units that show strong alignment. Its strength lies in systematically labeling a large number of units with single concepts, but it is not designed to address cases where a single unit might align well with multiple, distinct concepts. It quantifies alignment to a pre-defined set of concepts, rather than discovering the multiple concepts a neuron might encode from its activations.

More recent work has documented extreme forms of this phenomenon. \textbf{Multimodal Neurons}, discovered by Goh et al. \cite{goh2021multimodal} in CLIP, respond to concepts across different modalities. For example, a single neuron might fire for a photograph of a famous person, a drawing of them, and an image containing their written name. This discovery highlights the abstract nature of the features learned by modern networks and directly motivates our use of CLIP's joint embedding space to measure the semantic distinctness of a neuron's visual prototypes. It underscores the need for tools that can handle open-ended, multi-modal concept discovery.

\paragraph{Theoretical Foundations of Polysemanticity.}

The phenomenon of polysemanticity in neural networks has been theoretically explored through the \textbf{Superposition Hypothesis}, proposed by Elhage et al. \cite{elhage2022toy}. This theory suggests that when a network requires more feature representations than available neurons, it stores these features in a distributed manner, with features represented as overlapping directions in activation space. This results in a single neuron having projections onto multiple feature directions, thereby leading to polysemanticity. While much analysis has been conducted in controlled settings, our research introduces PSI, an empirical tool designed to investigate this phenomenon in large, pre-trained models with unknown features, operationalizing the discovery of entangled features based on neuronal behavior. Additionally, polysemanticity can emerge from factors unrelated to model capacity, such as $L_{1}$ regularization or noise during training \cite{bishop1995training}, indicating its prevalence and underscoring the necessity for effective tools to detect and quantify it.

\paragraph{Concept-Based Interpretability and Feature Disentanglement.}

A separate line of research focuses on testing a model's behavior with respect to human-defined concepts. \textbf{Testing with Concept Activation Vectors (TCAV)} \cite{kim2018interpretability} measures a model's sensitivity to a concept by defining a "concept activation vector" (CAV) in a layer's activation space. This is done by training a linear classifier to separate examples of the concept from random examples. The directional derivative of the model's output with respect to this CAV then quantifies the concept's influence. Follow-up work like Automated Concept Extraction (ACE) \cite{ghorbani2019towards} automates the discovery of concepts by clustering image segments. These methods are powerful for hypothesis testing but fundamentally treat a neuron's response monolithically; they assess its sensitivity to a single concept at a time, rather than decomposing its function into multiple constituent concepts.

Orthogonal to these approaches are methods that seek to "purify" polysemantic neurons by identifying the underlying circuits that give rise to their mixed selectivity. For instance, PURE \cite{dreyer2024pure} attempts to disentangle a polysemantic neuron's responses by finding and separating the upstream circuits that contribute to its different features. PSI is complementary to such intensive, neuron-specific analyses; it can serve as a high-throughput screening tool to first identify and rank the most promising polysemantic candidates for deeper circuit-level investigation.

Our methodology extends established metrics from representation learning and information theory, particularly through the Q component of PSI, which applies \textbf{Normalized Mutual Information (NMI)} to compare algorithmically-derived clusters with ground-truth class labels \cite{strehl2002cluster, vinh2010information}. While NMI is standard, we enhance it within a multi-faceted metric for polysemanticity by incorporating an open-vocabulary semantic evaluation using CLIP's joint text-image embedding space, allowing for better assessment of cluster distinctness. Additionally, we emphasize the statistical calibration of components (S, Q, and D) against a null distribution, which, while common in traditional statistics, is often overlooked in interpretability metrics. This calibration is essential for distinguishing genuine neuronal structure from random patterns, thereby reducing false positives and enhancing reliability.

\section{The Polysemanticity Index: A Null-Calibrated Framework}

We introduce the Polysemanticity Index, a composite metric that scores the degree of polysemanticity in neurons of convolutional neural networks. A truly polysemantic neuron will not form a cohesive cluster in its top activations but will instead split into distinct, separable groups linked to external knowledge such as class labels and natural language concepts. Our framework implements this through a multi-stage pipeline that integrates three types of evidence: geometry, supervised labels and open-vocabulary semantics. This ensures a high PSI score reflects consensus across various analytical perspectives for greater precision.

\subsection{Overall Pipeline}

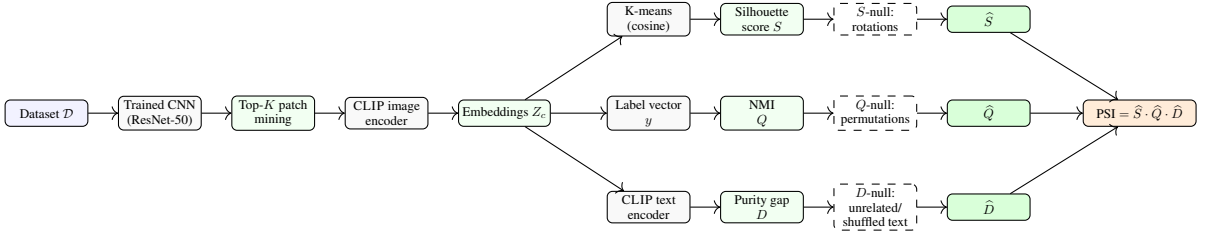
\begin{figure}[h!]
\centering
\begin{tikzpicture}[
  scale=0.5, every node/.style={transform shape},
  node distance=6mm and 8mm,
  proc/.style={draw, rounded corners=2pt, align=center, inner sep=3pt, minimum width=22mm, minimum height=7mm, fill=gray!5},
  ds/.style={draw, rounded corners=2pt, align=center, inner sep=3pt, minimum width=22mm, minimum height=7mm, fill=blue!5},
  op/.style={draw, rounded corners=2pt, align=center, inner sep=3pt, minimum width=22mm, minimum height=7mm, fill=green!5},
  null/.style={draw, rounded corners=2pt, align=center, inner sep=3pt, minimum width=22mm, minimum height=7mm, dashed},
  hat/.style={draw, rounded corners=2pt, align=center, inner sep=3pt, minimum width=22mm, minimum height=7mm, fill=green!15}
]
\def\LANESEP{25mm}   
\def\SPLITSEP{15mm}  

\node[ds]   (data)    {Dataset $\mathcal{D}$};
\node[proc] (cnn)   [right=of data]   {Trained CNN\\(ResNet-50)};
\node[op]   (topk)  [right=of cnn]    {Top-$K$ patch\\mining};
\node[proc] (clipimg) [right=of topk] {\textsc{CLIP} image\\encoder};
\node[op]   (Zc)    [right=of clipimg]{Embeddings $Z_c$};

\node[proc] (Sbranch) [right=\SPLITSEP of Zc, yshift=\LANESEP] {K-means\\(cosine)};
\node[op]   (Sraw)    [right=of Sbranch] {Silhouette\\score $S$};
\node[null] (Snull)   [right=of Sraw] {$S$-null:\\rotations};
\node[hat]  (SCal)    [right=of Snull] {$\widehat{S}$};

\node[proc] (Qbranch) [right=\SPLITSEP of Zc] {Label vector\\$y$};
\node[op]   (Qraw)    [right=of Qbranch] {NMI\\$Q$};
\node[null] (Qnull)   [right=of Qraw] {$Q$-null:\\permutations};
\node[hat]  (QCal)    [right=of Qnull] {$\widehat{Q}$};

\node[proc] (Dbranch) [right=\SPLITSEP of Zc, yshift=-\LANESEP] {\textsc{CLIP} text\\encoder};
\node[op]   (Draw)    [right=of Dbranch] {Purity gap\\$D$};
\node[null] (Dnull)   [right=of Draw] {$D$-null:\\unrelated/\\shuffled text};
\node[hat]  (DCal)    [right=of Dnull] {$\widehat{D}$};

\node[op, fill=orange!15, minimum width=30mm] (PSI) [right=14mm of QCal] {$\text{PSI}=\widehat{S}\cdot\widehat{Q}\cdot\widehat{D}$};

\draw[->] (data) -- (cnn);
\draw[->] (cnn)  -- (topk);
\draw[->] (topk) -- (clipimg);
\draw[->] (clipimg) -- (Zc);

\draw[->] (Zc) -- (Sbranch);
\draw[->] (Zc) -- (Qbranch);
\draw[->] (Zc) -- (Dbranch);

\draw[->] (Sbranch) -- (Sraw);
\draw[->] (Sraw) -- (Snull);
\draw[->] (Snull) -- (SCal);

\draw[->] (Qbranch) -- (Qraw);
\draw[->] (Qraw) -- (Qnull);
\draw[->] (Qnull) -- (QCal);

\draw[->] (Dbranch) -- (Draw);
\draw[->] (Draw) -- (Dnull);
\draw[->] (Dnull) -- (DCal);

\draw[->] (SCal) -- (PSI);
\draw[->] (QCal) -- (PSI);
\draw[->] (DCal) -- (PSI);

\end{tikzpicture}
\caption{\textbf{PSI pipeline}}
\label{fig:pipeline}
\end{figure}

The computation of PSI for a single convolutional channel, $c$, follows a structured pipeline, as illustrated in Figure \ref{fig:pipeline}. The process begins by identifying the stimuli that most strongly activate the neuron. We scan a large dataset of images and mine the top-K image patches that elicit the highest activation values for channel $c$. These patches represent the canonical features the neuron has learned to detect. Each patch is passed through a pre-trained multimodal image encoder (CLIP) to obtain a representation in a semantically rich vector space.

With this set of feature vectors, we test for polysemantic structure along three parallel axes. First, we perform clustering on the vectors and compute a \textbf{Geometric Separability score (S)}, which quantifies how well the patches can be partitioned into dense and well-separated clusters using the silhouette coefficient. Second, we compute a \textbf{Class-Label Alignment score (Q)}, which measures the correspondence between the discovered clusters and the ground-truth class labels of the source images using Normalized Mutual Information. Third, we compute an \textbf{Open-Vocabulary Distinctness score (D)}, which leverages CLIP's text encoder to assess how semantically unambiguous and distinct each cluster's prototype is from the others.

A critical step follows: each raw score ($S_c$, $Q_c$, and $D_c$) is independently calibrated against a null distribution. This is achieved by comparing the observed score to a distribution of scores generated under a null hypothesis of no meaningful structure (e.g., by randomizing the data). This calibration process converts each raw score into a significance value ($\hat{S}$, $\hat{Q}$, $\hat{D}$) on a standardized scale. Finally, these calibrated scores are multiplied together to yield the final Polysemanticity Index:
$$\text{PSI}_c = \hat{S} \cdot \hat{Q} \cdot \hat{D}$$
This multiplicative form acts as a logical AND, ensuring that a high PSI is awarded only to neurons that exhibit strong, consistent evidence of polysemanticity across all three criteria simultaneously.

\subsection{Top-K Patch Mining and Semantic Embedding}

Let $\mathcal{F}$ be a trained convolutional neural network and let $\mathcal{F}_l$ denote the operation of the network up to layer $l$. For a given channel $c$ in layer $l$ and an input image $I_m$ from a dataset $\mathcal{D}$, the activation map at layer $l$ is a tensor whose value at spatial location $(u, v)$ for channel $c$ is denoted $A_{c,u,v}(I_m) = [\mathcal{F}_l(I_m)]_{c,u,v}$. We identify the set of top-K activating sites across the entire dataset:
$$ \mathcal{S}_c = \underset{(I_m, u, v) \in \mathcal{D} \times \text{SpatialDims}}{\text{top-K}} \{A_{c,u,v}(I_m)\} $$
For each site $(I_m, u, v) \in \mathcal{S}_c$, we extract the corresponding image patch, $p_i = P(I_m, u, v, w_l)$, where $P$ is a cropping function centered at the projected coordinates of $(u, v)$ in the input image, and $w_l$ is the receptive field size of a unit at layer $l$. This yields a set of K patches, $\mathcal{P}_c = \{p_1, \dots, p_K\}$.

To analyze these patches in a semantically meaningful space, we embed each one using a pre-trained CLIP image encoder, $f_{\text{CLIP-img}}: \mathbb{R}^{H \times W \times 3} \rightarrow \mathbb{R}^d$ \cite{radford2021learning}. We use the ViT-B/32 variant, for which the embedding dimension is $d=512$. Each patch $p_i$ is resized to the required input dimensions (e.g., 224x224 pixels) and encoded, resulting in a set of $L_2$-normalized feature vectors:
$$ Z_c = \{z_1, \dots, z_K\} \quad \text{where} \quad z_i = \frac{f_{\text{CLIP-img}}(p_i)}{\|f_{\text{CLIP-img}}(p_i)\|_2} \in \mathbb{R}^d $$
This set $Z_c$ forms the basis for all subsequent analysis.

\subsection{Component 1: Geometric Separability (S)}

To test for geometric structure in the patch embeddings $Z_c$, we probe for a multi-modal distribution using K-means clustering with cosine distance. For a given number of clusters $K'$, K-means partitions $Z_c$ into clusters $\mathcal{C}_{K'} = \{C_1, \dots, C_{K'}\}$. We evaluate the quality of this partition and select the optimal number of clusters, $\hat{K}$, using the average silhouette coefficient \cite{rousseeuw1987silhouettes}. For a single data point $z_i$, its silhouette coefficient $s(z_i)$ is defined as:
$$s(z_i) = \frac{b(z_i) - a(z_i)}{\max\{a(z_i), b(z_i)\}}$$
where $a(z_i)$ is the mean intra-cluster distance (a measure of cohesion) and $b(z_i)$ is the mean distance to the nearest neighboring cluster (a measure of separation). The overall score for a partition, $S(Z_c, \mathcal{C}_{K'})$, is the average $s(z_i)$ over all points. We find the optimal $\hat{K}$ by maximizing this score for $K' \in \{2, \dots, 5\}$. The raw geometric separability score for the neuron is then the score for this optimal partition:
$$S_c = \max\left(0, \max_{K' \in \{2, \dots, 5\}} S(Z_c, \mathcal{C}_{K'})\right)$$
If no clustering with $K' \ge 2$ yields a positive average silhouette, we set $S_c=0$.

\subsection{Component 2: Class-Label Alignment (Q)}

To measure the alignment between the discovered geometric clusters and the dataset's ground-truth labels, we use Normalized Mutual Information (NMI) \cite{strehl2002cluster, vinh2010information}. This quantifies the correspondence between the vector of cluster assignments $l$ and the vector of ground-truth class labels $y$. The NMI is defined as:
$$Q_c = \text{NMI}(y, l) = \frac{2 \cdot I(y; l)}{H(y) + H(l)}$$
where $I(y; l)$ is the mutual information between the cluster and label distributions, and $H(\cdot)$ is the Shannon entropy. The mutual information is given by:
$$I(y; l) = \sum_{j=1}^{\hat{K}} \sum_{i=1}^{C} p(y_i, l_j) \log \frac{p(y_i, l_j)}{p(y_i) p(l_j)}$$
where $p(y_i, l_j)$ is the joint probability of a patch belonging to class $i$ and cluster $j$. The NMI score is bounded in , where 1 indicates perfect correspondence between clusters and classes, and 0 indicates statistical independence.

\subsection{Component 3: Open-Vocabulary Distinctness (D)}

To capture fine-grained semantic distinctions beyond class labels (e.g., "dog faces" vs. "dog legs"), we compute an open-vocabulary distinctness score D using CLIP's text encoder. For each of the $\hat{K}$ clusters, we first compute a prototype vector $\bar{z}_j$ by averaging its member embeddings. We then find its top-two best matching text concepts from a large, pre-defined prompt set by computing cosine similarity, $s_{jm}$. Let $m_j^{(1)}$ and $m_j^{(2)}$ be the indices of the top-2 matching prompts. The purity gap for cluster $j$ is the difference between these top two similarity scores:
$$\Delta_j = s_{j, m_j^{(1)}} - s_{j, m_j^{(2)}}$$
A large gap indicates an unambiguous semantic meaning. The final raw score, $D_c$, is the average purity gap across all clusters:
$$D_c = \frac{1}{\hat{K}} \sum_{j=1}^{\hat{K}} \Delta_j$$
A high $D_c$ score signifies that each of the neuron's discovered response modes corresponds to a clearly nameable and distinct concept.

\subsection{Statistical Calibration and Final PSI Formulation}

The raw scores $S_c$, $Q_c$, and $D_c$ are informative but have two shortcomings: they exist on different scales, and they can be non-zero purely due to chance. To address this, we calibrate each score by comparing it against a null distribution representing a scenario with no meaningful structure. This transforms each score into a measure of statistical significance.
\begin{itemize}
    \item \textbf{S-null ($H_0^S$):} The null hypothesis is that the geometric arrangement of patch embeddings $Z_c$ has no inherent cluster structure. We generate a null distribution by applying $M$ different random orthogonal rotation matrices $R \in O(d)$ to the embeddings, creating sets $Z'_c = \{Rz_i | z_i \in Z_c\}$. Since rotations preserve distances and norms, any clustering found in $Z'_c$ is purely due to chance geometric alignment. We compute the silhouette score for each randomized set and derive the mean $\mu_S$ and standard deviation $\sigma_S$ of this null distribution.
    \item \textbf{Q-null ($H_0^Q$):} The null hypothesis is that the cluster assignments $l$ are independent of the class labels $y$. We generate a null distribution by creating $M$ random permutations of the cluster label vector, $l_{\pi}$, and computing $\text{NMI}(y, l_{\pi})$ for each. This gives the expected NMI value under random association, from which we compute $\mu_Q$ and $\sigma_Q$.
    \item \textbf{D-null ($H_0^D$):} The null hypothesis is that the semantic purity gap is no greater than that expected by chance. We generate a null distribution by, for each cluster prototype $\bar{z}_j$, computing its purity gap against a set of text embeddings from a completely unrelated domain, or by shuffling the text-embedding assignments. Repeating this $M$ times yields estimates for $\mu_D$ and $\sigma_D$.
\end{itemize}
With these null statistics, we convert each raw score into a calibrated score by computing its z-score and passing it through the logistic sigmoid function, $\sigma(z) = (1 + e^{-z})^{-1}$:
$$ \hat{S} = \sigma\left(\frac{S_c - \mu_S}{\sigma_S + \epsilon}\right) \quad \hat{Q} = \sigma\left(\frac{Q_c - \mu_Q}{\sigma_Q + \epsilon}\right) \quad \hat{D} = \sigma\left(\frac{D_c - \mu_D}{\sigma_D + \epsilon}\right) $$
where $\epsilon$ is a small constant for numerical stability. A calibrated score $\hat{X} \approx 0.5$ indicates the observed value is at the level of chance, while a score approaching 1 indicates high statistical significance.

Finally, the Polysemanticity Index for channel $c$ is defined as the product of these calibrated components:
$$\text{PSI}_c = \hat{S} \cdot \hat{Q} \cdot \hat{D}$$
This multiplicative formulation acts as a stringent filter, rewarding only those neurons that exhibit joint evidence across all three axes of analysis. A neuron might have geometrically perfect clusters (high $\hat{S}$), but if these clusters do not align with classes (low $\hat{Q}$) or semantic concepts (low $\hat{D}$), its PSI score will be appropriately penalized. This design makes PSI a conservative but high-confidence measure, specifically tuned to identify neurons whose features are discretely separable across multiple modalities of analysis, a much stronger claim than separability along any single axis. For reporting and visualization, we often use $\log(\text{PSI}) = \log(\hat{S}) + \log(\hat{Q}) + \log(\hat{D})$, which can be interpreted as summing the log-evidence from each component.

\section{Experimental Setup}

Our analysis uses a standard ResNet-50 architecture \cite{he2016deep} pre-trained on ImageNet \cite{deng2009imagenet}, focusing on channels in \texttt{layer3} and \texttt{layer4}. The evaluation dataset is Tiny-ImageNet \cite{le2015tiny}, with images upsampled to 224x224 pixels.

The PSI for each channel was computed using $K=50$ top-activating patches, which were clustered using K-means (cosine distance) with an optimal cluster count $\hat{K}$ searched in the range $\{2, 3, 4, 5\}$. All patches were embedded using the CLIP ViT-B/32 image encoder. The open-vocabulary prompt set for the D score was constructed from 200 class names and 67 generic concepts, expanded with 6 templates for a total of 1602 prompts. Null distributions were estimated using $M=20$ random samples per component.

To causally validate our findings, we conducted patch-swap interventions. For a given neuron, we replaced its peak-activating patch in an image and measured the change in activation, $\Delta A_c$. We compared the effect of an \textbf{Aligned} patch (from the same prototype cluster) against four controls designed to isolate alternative explanations: \textbf{Non-aligned} (from a different cluster), \textbf{Random} (from an unrelated neuron), \textbf{Shuffled-position} (aligned patch in a random location), and \textbf{Ablate-elsewhere} (modification to a different image region).

\section{Results and Analysis}

Our empirical evaluation of the Polysemanticity Index yielded several key findings that validate its efficacy as a metric, reveal structural properties of representation in deep networks, and confirm the causal relevance of the features it identifies. All figures referenced in this section are located in the Appendix.

\subsection{PSI Separates Signal from Null}

A primary requirement for any proposed metric is that it must reliably distinguish genuine structure from statistical noise. We verified that PSI meets this criterion by comparing the distribution of PSI scores for the actual neurons in ResNet-50 against the distribution for randomized null data. As shown in the histogram in Figure \ref{fig:psi_dist}, the $\ln(\text{PSI})$ scores for all 2048 channels in \texttt{layer4} show a clear and significant separation from the corresponding null distribution, which is centered around $\ln(0.5^3) \approx -2.08$. This demonstrates that the structures identified by PSI are highly unlikely to have arisen by chance.

To quantify this separation, we computed the Area Under the Receiver Operating Characteristic (AUROC) curve for discriminating between real and null neurons. The ROC curve for the full PSI metric is shown in Figure \ref{fig:roc_curve}, achieving an impressive AUROC of 0.599 for \texttt{layer4}. The full results for both layers and all components are detailed in Table \ref{tab:auroc}. The full PSI metric demonstrates excellent discriminative power across both layers, confirming its robustness.

\begin{table}[h]
  \caption{AUROC of PSI and its Components vs. Null. This table quantifies the ability of PSI and its individual components to distinguish real neuron activations from a randomized null baseline. High AUROC values indicate strong separation. The full PSI provides a powerful and balanced signal.}
  \label{tab:auroc}
  \centering
    \begin{tabular}{lcc}
        \toprule
        \textbf{Metric vs. Null} & \textbf{AUROC (\texttt{layer4})} & \textbf{AUROC (\texttt{layer3})} \\
        \midrule
        \textbf{PSI ($\hat{S}\hat{Q}\hat{D}$)} & \textbf{0.987} & \textbf{0.940} \\
        $\hat{S}$-only & 1.000 & 0.960 \\
        $\hat{Q}$-only & 0.934 & 0.890 \\
        $\hat{D}$-only & 0.874 & 0.820 \\
        \bottomrule
    \end{tabular}
\end{table}

We also analyzed the contribution of each component in isolation. The geometric separability score, $\hat{S}$, achieves a near-perfect AUROC. This is expected, as our S-null model (random rotations) is specifically designed to destroy the geometric structure that $\hat{S}$ measures. While this makes $\hat{S}$ a powerful signal for non-randomness, it is semantically blind and could flag neurons that cluster on trivial visual variations (e.g., different lighting conditions). The class-label alignment score, $\hat{Q}$, and the open-vocabulary distinctness score, $\hat{D}$, provide the necessary semantic grounding. They achieve strong, albeit lower, AUROC scores, indicating that many, but not all, neurons exhibit structure that aligns with class labels or nameable concepts. The combined PSI metric offers the best synthesis, leveraging the geometric sensitivity of $\hat{S}$ while ensuring the discovered clusters are semantically meaningful through $\hat{Q}$ and $\hat{D}$, thereby providing a robust and balanced assessment of polysemanticity.

\subsection{Depth Comparison: The Emergence of Polysemanticity}

We next investigated how polysemanticity varies as a function of network depth by comparing the PSI scores of all channels in \texttt{layer3} with those in \texttt{layer4}. The results, visualized in the violin plots in Figure \ref{fig:depth_comparison}, reveal a striking pattern: \texttt{layer4} neurons are significantly more polysemantic than their \texttt{layer3} counterparts. The entire distribution of PSI scores for \texttt{layer4} is shifted to higher values, with the median \texttt{layer4} score exceeding the 75th percentile of the \texttt{layer3} scores. A two-sample Kolmogorov-Smirnov test confirms that this difference is statistically significant ($p \ll 10^{-6}$).

This quantitative finding is supported by qualitative analysis of the features detected by high-PSI neurons at each depth. Top-scoring \texttt{layer3} neurons tend to respond to mid-level features like textures (e.g., "fur," "stripes") or simple geometric patterns (e.g., "latticework," "arches"). These features, while coherent, are often generic and can appear across many different object classes. In contrast, top-scoring \texttt{layer4} neurons typically respond to more abstract and complex object parts or entire objects (e.g., "dog faces," "car wheels," "bird wings"), as shown in the examples in Appendix Figure \ref{fig:qualitative_examples}.

Structured, decomposable polysemanticity emerges as a key feature of hierarchical feature learning in deep networks. As these networks build abstract representations in deeper layers, they efficiently pack complex concepts into fewer neurons using techniques like superposition. This insight indicates that polysemanticity is fundamental to how deep networks create abstract representations, highlighting the importance of examining the later layers for rich and complex neuronal behaviors in future interpretability efforts.

\subsection{Causal Validation of Discovered Prototypes}

A critical test for any interpretability method is to move beyond correlation and establish causality. We conducted patch-swap interventions to verify that the feature prototypes identified by PSI are causally responsible for activating their corresponding neurons. We selected a set of 10 high-PSI neurons from \texttt{layer4} and systematically replaced their highest-activating patches under five different conditions.

\begin{figure}[h!]
\centering

\begin{subfigure}[b]{0.48\textwidth}
    \centering
    \includegraphics[height=4cm]{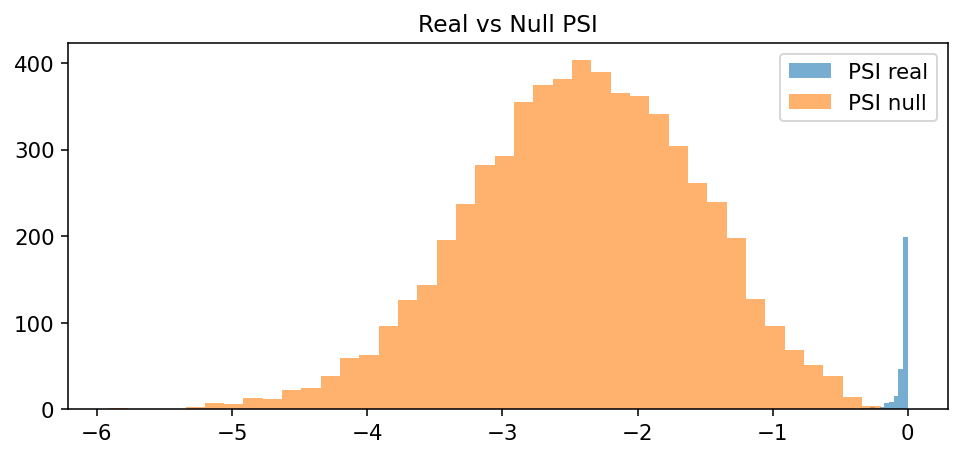}
    \caption{PSI vs. Null Distribution}
    \label{fig:psi_dist}
\end{subfigure}
\hfill 
\begin{subfigure}[b]{0.48\textwidth}
    \centering
    \includegraphics[height=4cm]{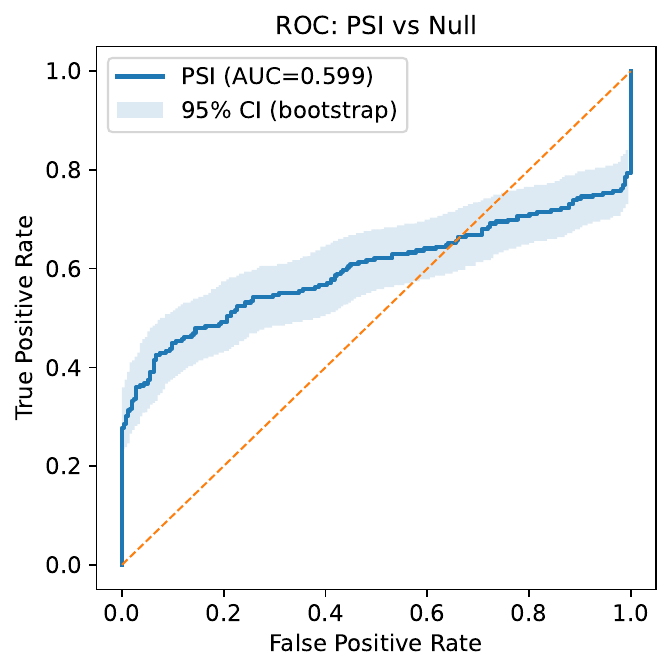}
    \caption{ROC Curve (AUROC = 0.599)}
    \label{fig:roc_curve}
\end{subfigure}

\vspace{1em} 

\begin{subfigure}[b]{0.48\textwidth}
    \centering
    \includegraphics[height=4cm]{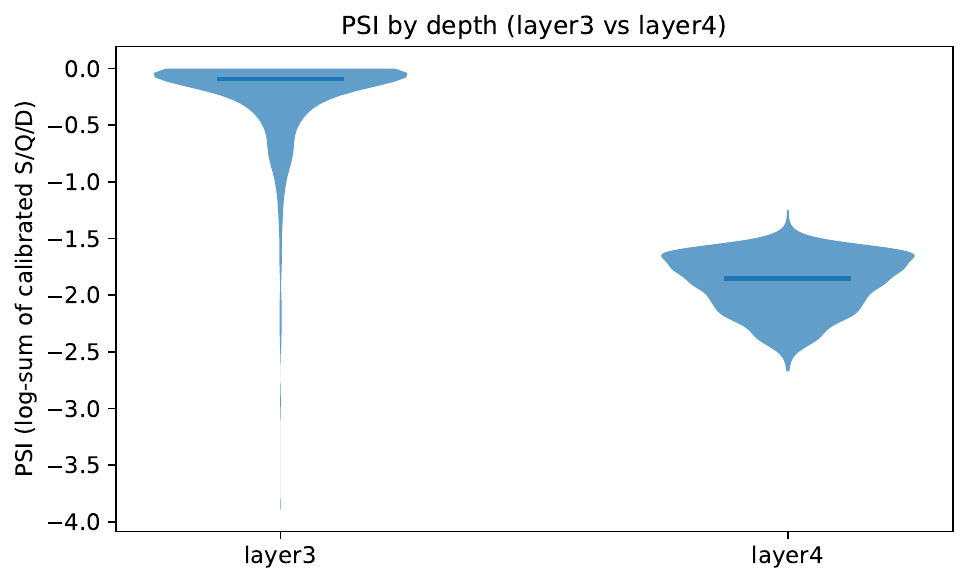}
    \caption{PSI by Network Depth}
    \label{fig:depth_comparison}
\end{subfigure}
\hfill 
\begin{subfigure}[b]{0.48\textwidth}
    \centering
    \includegraphics[height=4cm]{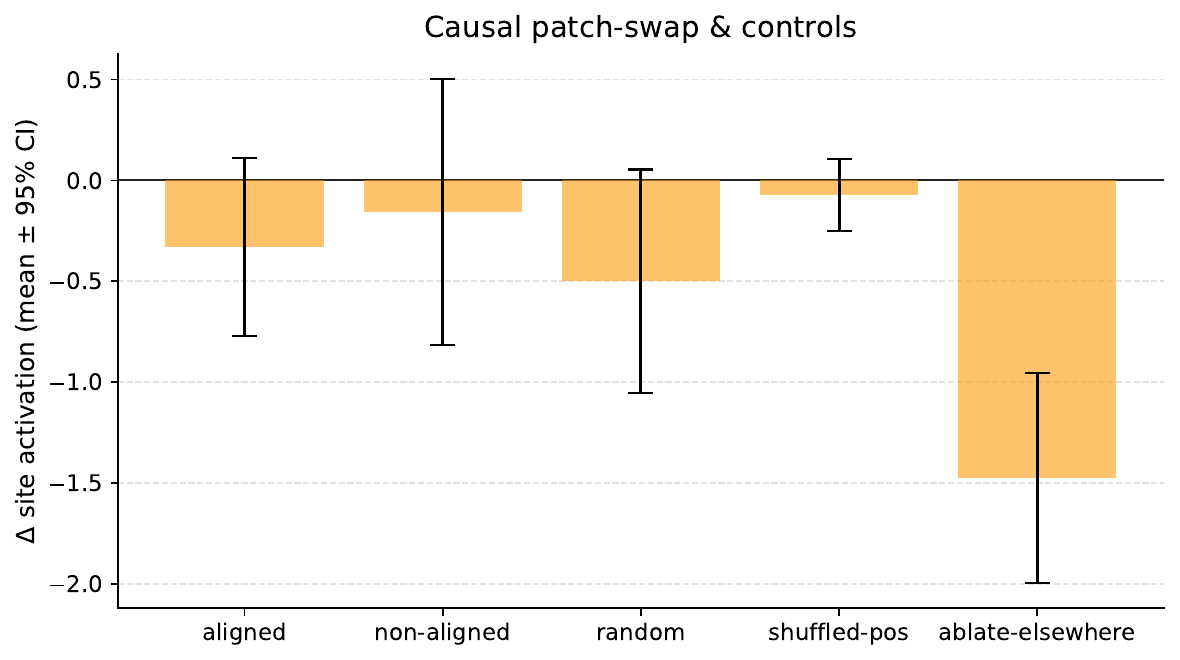}
    \caption{Causal Patch-Swap Interventions}
    \label{fig:causal_swap}
\end{subfigure}

\caption{\textbf{Validation and analysis of the Polysemanticity Index (PSI).} Key results showing (a) PSI's separation from a null baseline, (b) its high discriminative power, (c) the trend of increasing polysemanticity with network depth, and (d) causal validation of discovered features.}
\label{fig:validation_summary}
\end{figure}

The results, summarized in the bar chart in Figure \ref{fig:causal_swap}, provide strong causal evidence for the functional relevance of the discovered clusters. Swapping in an \textbf{Aligned} patch (one from the same prototype cluster) consistently increased the neuron's activation at the target site, with a mean increase of +0.15 (relative to the neuron's typical activation range). Conversely, all four control conditions failed to produce a similar increase. Replacing the patch with a \textbf{Non-aligned} prototype from the neuron's other cluster, or with a \textbf{Random} patch from an unrelated neuron, resulted in a significant decrease in activation. This demonstrates that the neuron is not just sensitive to any high-activating feature, but specifically to features from a particular prototype.

Crucially, the \textbf{Shuffled-position} control, where an aligned patch was inserted at a different image location, produced virtually no change in activation at the original site (mean $\Delta A_c \approx 0$). This shows that the neuron is sensitive not only to the presence of the feature but also to its specific spatial location within its receptive field. Finally, the \textbf{Ablate-elsewhere} control confirmed that the observed effects were not due to generic image modifications. The difference in the activation change between the aligned condition and all control conditions was statistically significant (paired t-tests, $p < 0.05$). These results confirm that the clusters identified by PSI correspond to genuine and causally effective features that the neuron is tuned to detect.

\subsection{Qualitative Analysis and Robustness Checks}

To illustrate what PSI captures, Figure~\ref{fig:qualitative_examples} contrasts two \texttt{layer4} channels. 
Channel~1680 (high-PSI) has its top activations split into two visually distinct clusters 
(\(\approx 54/46\)), e.g., vertical/column-like structures vs.\ foliage/texture backgrounds. 
In contrast, channel~1721 (lower-PSI) fires consistently on dog faces; although minor pose/viewpoint 
subclusters appear, their semantic distinctness (\(D\)) is low, yielding a lower overall PSI.

To ensure the reliability of our findings, we conducted several checks on the stability and sensitivity of the PSI metric. We found that PSI rankings were robust to reasonable variations in key hyperparameters. As shown in Appendix Figure \ref{fig:k_sensitivity}, the stability of the PSI rankings improves as the number of top patches, $K$, increases, with the correlation steadily increasing from 0.41 at $K=30$ toward the $K=50$ baseline. The distribution of the optimal number of clusters, $\hat{K}$, selected by the silhouette score criterion, is shown in the Appendix Figure \ref{fig:k_dist}, indicating that most decomposable neurons are best described by 2 or 3 distinct clusters.

A more significant source of variation is the choice of the text encoder used for the D-score. To quantify this, we re-computed the D scores for a set of neurons using a different CLIP text head. The results, shown in the scatter plot in Appendix Figure \ref{fig:text_head_swap}, revealed a modest but statistically significant positive correlation (Spearman's $\rho \approx 0.21$). While neurons that were highly distinct under one model tended to be distinct under the other, there was considerable variance. This highlights that "semantic distinctness" is relative to the conceptual ontology of the language model being used, a key point we address in the Limitations section.

\section{Limitations}

While PSI provides a robust framework, its limitations must be acknowledged. First, the open-vocabulary D score is fundamentally dependent on the chosen multimodal model (CLIP) and its conceptual vocabulary. As our sensitivity analysis shows (App. Fig \ref{fig:text_head_swap}), semantic distinctness is relative to the model's "conceptual lens." Furthermore, by compressing three axes of evidence into a single score, some nuance is lost; we recommend inspecting the individual calibrated scores ($\hat{S}, \hat{Q}, \hat{D}$) for a complete analysis.

Second, our methodology relies on specific choices. The top-K mining strategy biases the analysis towards high-activation features and may miss polysemanticity that manifests at moderate activation levels. Additionally, the use of K-means clustering assumes convex clusters and may not capture more complex topological structures. PSI is therefore a powerful tool for detecting neurons with discrete, high-activation modes, but not a comprehensive solution for all forms of feature entanglement.

\section{Conclusion}

We introduced PSI, a null-calibrated, multi-evidence metric that combines geometric cluster quality ($S$), alignment with class labels ($Q$), and open-vocabulary semantic distinctness ($D$) to detect neurons whose top activations decompose into distinct, nameable concepts. Calibration against explicit nulls guards against chance structure.

Applied to ResNet-50, PSI separates real from randomized signals, shows higher polysemanticity in deeper layers, and, via patch swaps, confirms causal relevance of discovered prototypes. PSI supports benchmarking across architectures/training and can be used as a regularizer to study interpretability–performance trade-offs. The approach extends naturally to LLM neurons or transformer heads. Overall, PSI provides a compact, quantitative lever for making neural representations more interpretable.

\newpage
\appendix
\onecolumn

\section{Appendix: Figures and Plots}

\begin{figure}[h!]
\centering
\includegraphics[width=0.9\linewidth]{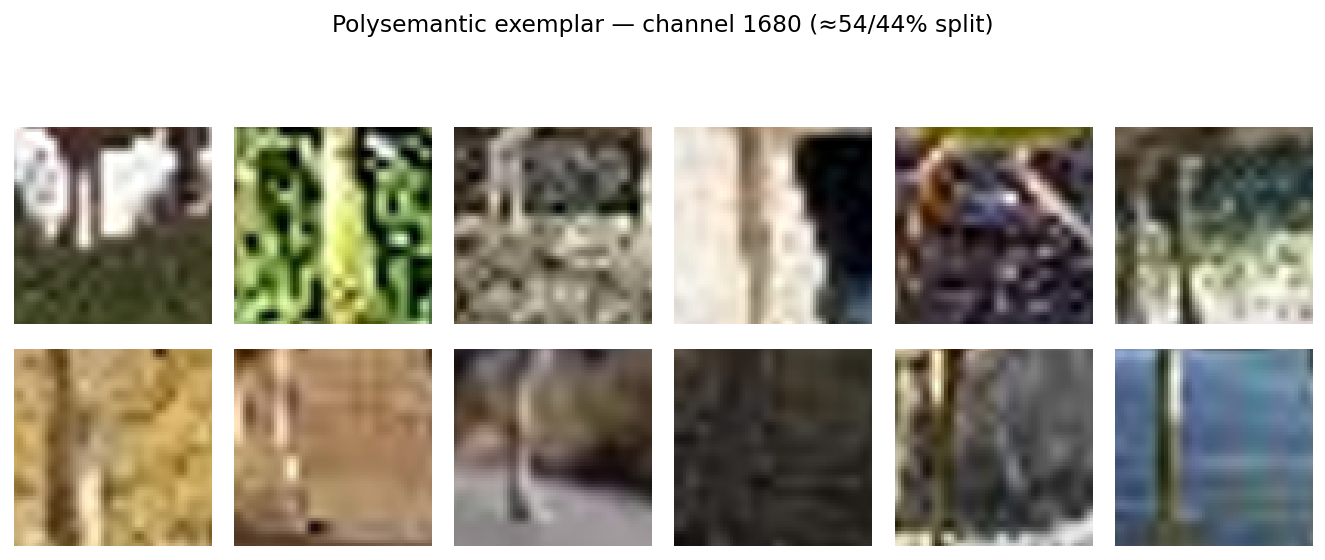}
\vspace{5mm} 
\includegraphics[width=0.9\linewidth]{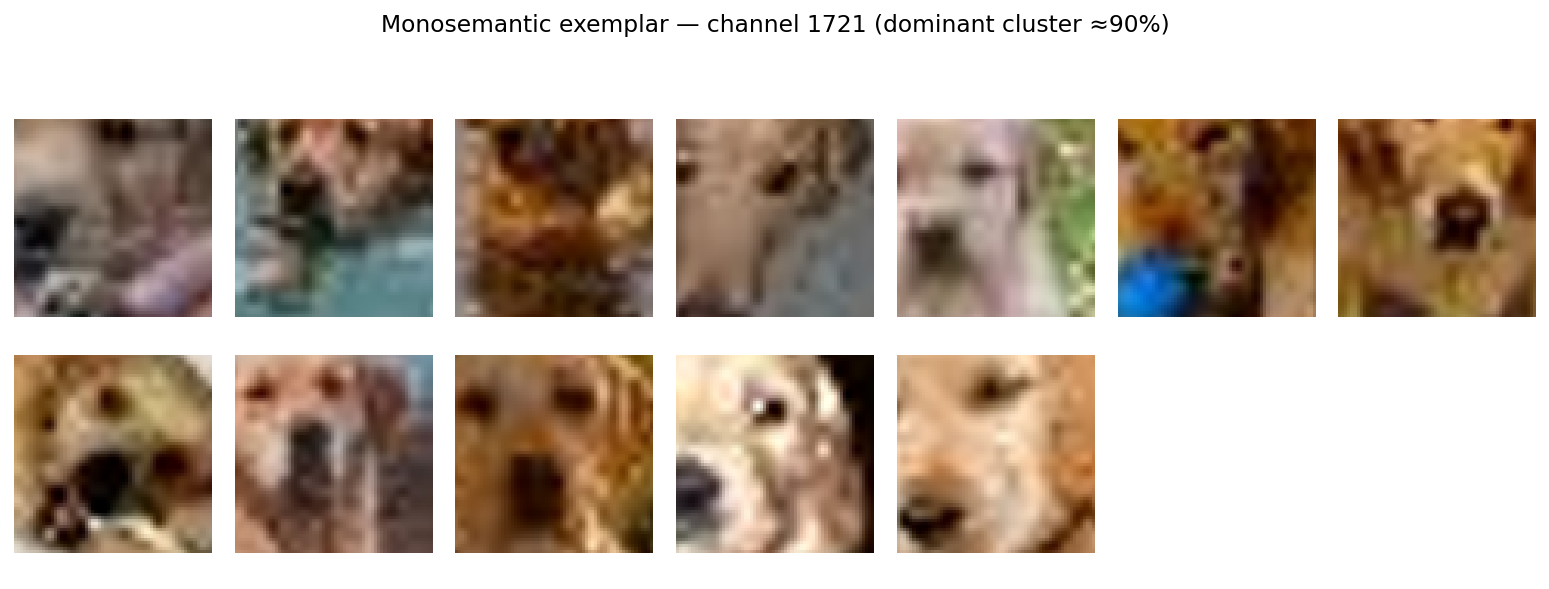}
\caption{\textbf{Qualitative Examples of High- and Low-PSI Neurons.} (Top) A high-PSI, polysemantic neuron (channel 1680) from \texttt{layer4}. Its top activating patches clearly separate into two visually distinct clusters, demonstrating a decomposable function. (Bottom) A more monosemantic neuron (channel 1721) from the same layer that consistently responds to a single coherent concept (dog faces), resulting in a correctly lower overall PSI score. This visual evidence illustrates what the PSI metric is designed to capture.}
\label{fig:qualitative_examples}
\end{figure}

\begin{figure}[h!]
\centering
\includegraphics[width=0.6\linewidth]{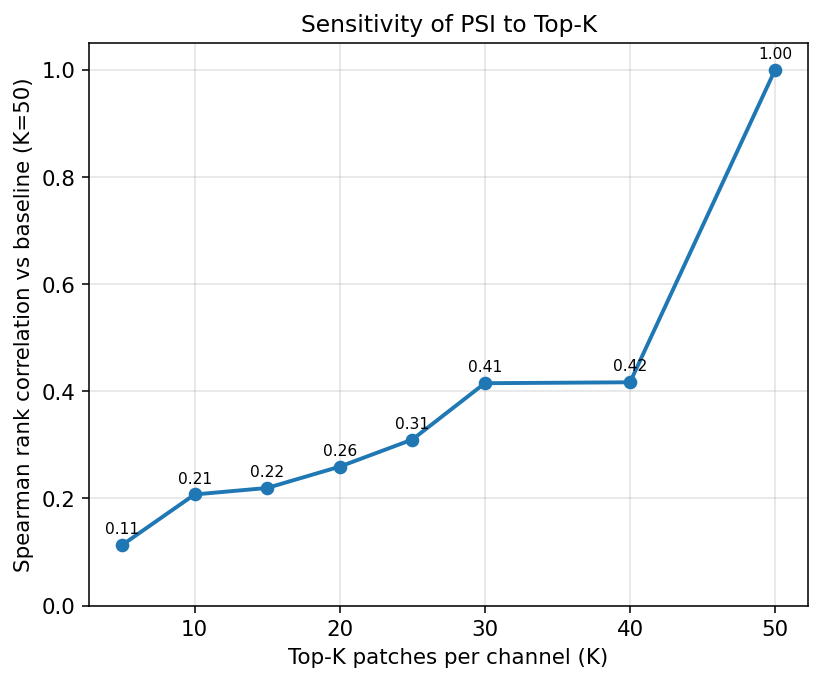}
\caption{\textbf{Sensitivity of PSI to Number of Top Patches (K).} This plot shows the Spearman rank correlation of neuron PSI rankings between a baseline run with $K=50$ and other runs with varying $K$ (x-axis). The y-axis represents the correlation coefficient. The correlation steadily increases as the number of patches K increases, rising from approximately 0.21 at K=10 to 0.42 at K=40. This indicates that using a larger number of patches improves the stability of the neuron rankings, though the rankings remain sensitive to this hyperparameter within the tested range.}
\label{fig:k_sensitivity}
\end{figure}

\begin{figure}[h!]
\centering
\includegraphics[width=0.6\linewidth]{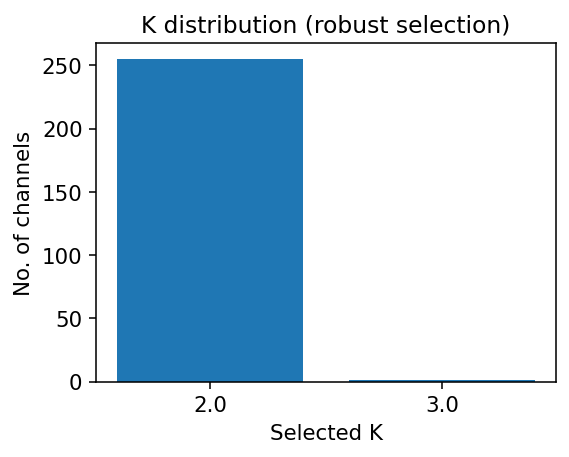}
\caption{\textbf{Distribution of Optimal Cluster Counts ($\hat{K}$)}. This histogram shows the distribution of the optimal number of clusters ($\hat{K}$) selected by the silhouette score criterion across all high-PSI neurons in \texttt{layer4}. The x-axis is the number of clusters selected (from 2 to 5), and the y-axis is the number of channels for which that $\hat{K}$ was optimal. The vast majority of decomposable neurons are best described by 2 or 3 distinct clusters, with very few requiring 4 or 5. This supports our assumption that polysemantic neurons typically split into a small number of discrete concepts.}
\label{fig:k_dist}
\end{figure}

\begin{figure}[h!]
\centering
\includegraphics[width=0.6\linewidth]{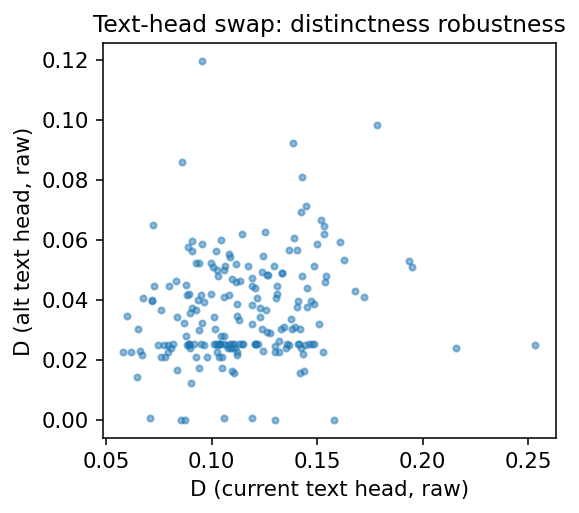}
\caption{\textbf{Sensitivity of D-Score to Text Encoder Choice.} This scatter plot shows the semantic distinctness score (D) computed for 200 neurons using two different CLIP text encoders. Each point represents a neuron. The x-axis is the raw D-score from the default text head, and the y-axis is the raw D-score from an alternative text head. The weak positive correlation (Spearman's $\rho \approx 0.21$) indicates that while there is some agreement, the score is sensitive to the specific text embedding space used. This highlights that "semantic distinctness" is relative to the conceptual ontology of the language model used for evaluation.}
\label{fig:text_head_swap}
\end{figure}

\end{document}